\documentclass[10pt,twocolumn]{article}
\usepackage{iccv}
\usepackage{times}
\usepackage{epsfig}
\usepackage{graphicx}
\usepackage{amsmath}
\usepackage{amssymb}
\usepackage{multirow, makecell}
\usepackage{colortbl} 
\usepackage{xcolor}
\usepackage{array}
\definecolor{Gray}{gray}{0.94}
\definecolor{liGray}{gray}{0.5}
\definecolor{LightCyan}{rgb}{0.88,1,1}
\usepackage{mathrsfs}
\newlength\savewidth\newcommand\shline{\noalign{\global\savewidth\arrayrulewidth
  \global\arrayrulewidth 1pt}\hline\noalign{\global\arrayrulewidth\savewidth}}
\newcommand{\tablestyle}[2]{\setlength{\tabcolsep}{#1}\renewcommand{\arraystretch}{#2}\centering\footnotesize}
\usepackage{bbding}

\usepackage[pagebackref=true,breaklinks=true,colorlinks,bookmarks=false]{hyperref}

\iccvfinalcopy 

\def\iccvPaperID{1353} 

\begin{document}

\title{Few-shot Action Recognition with Captioning Foundation Models}

\author{
   Xiang Wang$^{1*}$
     \hspace{0.02cm} 
    Shiwei Zhang$^{2\dag}$
     \hspace{0.02cm} 
    Hangjie Yuan$^{3}$
     \hspace{0.02cm} 
   Yingya Zhang$^2$ 
     \hspace{0.02cm}
    Changxin Gao$^1$ 
     \\
    Deli Zhao$^2$
    \hspace{0.02cm}
     Nong Sang$^{1\dag}$\\
    $^1$Key Laboratory of Image Processing and Intelligent Control,\\  \hspace{-0.5cm} School of Artificial Intelligence and Automation, Huazhong University of Science and Technology\\
     \hspace{-0.5cm}  $^2$Alibaba Group  \hspace{.2in} $^3$Zhejiang University \\
{\tt\footnotesize \{wxiang,cgao,nsang\}@hust.edu.cn, \{zhangjin.zsw,yingya.zyy\}@alibaba-inc.com,} \\
{\tt\footnotesize hj.yuan@zju.edu.cn, zhaodeli@gmail.com}
}
\maketitle
\let\thefootnote\relax\footnotetext{$*$ Intern at Alibaba Group. \hspace{1mm} $\dag$ Corresponding authors. }

\maketitle
\ificcvfinal\thispagestyle{empty}\fi

\begin{abstract}

Transferring vision-language knowledge from pretrained multimodal foundation models to various downstream tasks is a promising direction.
However, most current few-shot action recognition methods are still limited to a single visual modality input due to the high cost of annotating additional textual descriptions. 
In this paper, we develop an effective plug-and-play framework called CapFSAR to exploit the knowledge of multimodal models without manually annotating text.
%
%
%
To be specific, we first utilize a captioning foundation model ({i.e.}, BLIP) to extract visual features and automatically generate associated captions for input videos.
Then, we apply a text encoder to the synthetic captions to obtain representative text embeddings.
Finally, a visual-text aggregation module based on Transformer is further designed to incorporate cross-modal spatio-temporal complementary information for reliable few-shot matching.
In this way, CapFSAR can benefit from powerful multimodal knowledge of pretrained foundation models, yielding more comprehensive classification in the low-shot regime.
Extensive experiments on multiple standard few-shot benchmarks demonstrate that the proposed CapFSAR performs favorably against
existing methods and achieves state-of-the-art performance.
The code will be made publicly available.

\end{abstract}

\section{Introduction}

Few-shot action recognition aims to learn a generalizable model that can recognize new classes with a limited amount of videos.
Due to the high cost of collecting and annotating large-scale datasets, researchers have begun to pay considerable attention to this task recently and proposed a range of corresponding customized algorithms~\cite{CMN,OTAM,TRX,HyRSM,huang2022compound}.
%
%
%
%

\begin{figure}[t]
  \centering
   \includegraphics[width=0.96\linewidth]{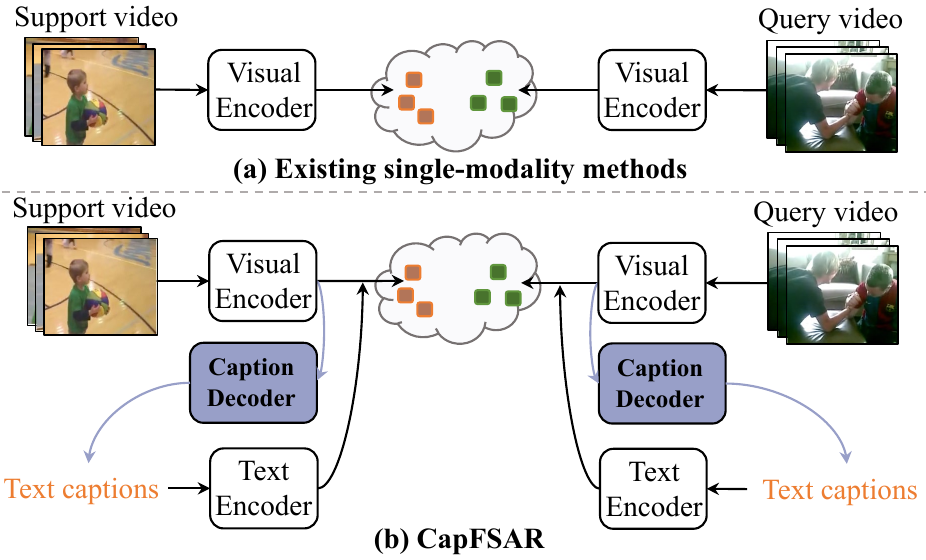}
\vspace{-1.5mm}
   \caption{
  Comparison between existing methods and the proposed CapFSAR.
  (a) Due to the high cost of manual text annotation, most existing techniques usually utilize a single visual modality for few-shot action recognition without involving multimodal information;
  (b) Our CapFSAR automatically generates additional captions for input videos to take advantage of the auxiliary text modality, yielding informative multimodal representations for few-shot matching.
}
   \label{fig:motivation}
   \vspace{-4mm}
\end{figure}

%
%
Recent attempts mainly focus on the metric-based meta-learning paradigm~\cite{MatchNet,prototypical} to learn a discriminative visual feature space for the input videos and employ a temporal metric for prediction.
Despite impressive progress, these methods are still struggling to natively use unimodal vision models without involving multimodal knowledge (Figure~\ref{fig:motivation}(a)), leading to insufficient information exploitation, especially in the data-limited condition. 
%

The current prevailing trend of transferring multimodal knowledge in vision-language pretraining models~\cite{CLIP,ALIGN,BLIP,wang2022ofa} to a diverse range of downstream tasks has been proven effective and achieved remarkable success~\cite{zhou2022learning,gu2021open,ju2022prompting,wang2022clip}.
A natural question arises: \textit{How can few-shot action recognition take advantage of the foundation model to mine the powerful multimodal knowledge?}
There are two intuitive alternatives to achieve this goal:
\textit{i}) Annotating additional texts for each input video, which appears to be time-consuming and expensive;
\textit{ii}) Constructing hand-crafted text prompts using the annotated action labels, which is intractable due to the inaccessible labels of the query video and high demands for professional domain knowledge (\eg, professional gymnastics).
Besides, there are possible scenes that are difficult to annotate action names manually and only contain non-descript task labels, \eg, tasks with numerical labels~\cite{wang2021role,vijayanarasimhan2010far}.
%
%
The aforementioned potential drawbacks seriously hinder the application of recent  multimodal foundation models in the few-shot action recognition field.
A possible solution to alleviate the above-mentioned labeling problems is to leverage existing captioning networks to automatically generate text descriptions for videos.
However, this idea relies on the quality of generated captions, which traditional captioning methods~\cite{caption_1,pan2017video,wang2018reconstruction,pei2019memory} usually cannot meet due to limited training data.
With the development of large-scale vision-language pretraining~\cite{CLIP,ALIGN,BLIP},
recent captioning foundation models such as BLIP~\cite{BLIP} achieve promising caption generation results through learning from hundreds of millions of image-text pairs and have been widely adopted in downstream applications~\cite{chen2022litevl,santurkar2022caption,wang2022position,vong2022few}.
Therefore, a natural thought is to leverage BLIP to automatically generate captions for videos.

Inspired by the collective observations above, we develop a simple yet effective framework, namely CapFSAR, which attempts to exploit multimodal knowledge to deal with information scarcity in few-shot conditions by automatically generating and utilizing textual descriptions via pretrained BLIP.
Concretely, as shown in Figure~\ref{fig:motivation}(b), we first utilize the visual encoder of BLIP to encode features for input videos. 
%
Then, a caption decoder is applied to synthesize captions, which can be considered as an auxiliary augmented view.
Subsequently, we treat the text descriptions as an interface to extract contextual knowledge of the text encoder.
Finally, we feed the obtained visual and text features into a Transformer-based visual-text aggregation module to perform spatio-temporal cross-modal interactions and further enhance the temporal awareness of the learned model.
By this means, the proposed CapFSAR can obtain rich complementary multimodal information from foundation models in data-limited scenarios. 
Extensive experimental results on five commonly used benchmarks demonstrate that  CapFSAR is powerful in learning multimodal features and consequently outperforms previous state-of-the-art methods. 

%
%

Our contributions can be summarized as follows: 
\textit{i}) To the best of our knowledge, CapFSAR is the first few-shot action recognition approach that automatically generates text descriptions and thus enables the exploitation of multimodal knowledge from foundation models.
We believe that CapFSAR will facilitate future research on using large-scale pretrained models;
\textit{ii}) We introduce a visual-text aggregation module to capture spatio-temporal complementary information for the input visual and text features;
\textit{iii}) Empirical results indicate that CapFSAR outperforms existing advanced methods and achieves state-of-the-art performance.

\section{Related Work}
We briefly review some related literature, including few-shot image classification, caption generation, multimodal foundation model, and few-shot action recognition.

\vspace{1mm}
\noindent \textbf{Few-shot image classification.}
Few-shot learning~\cite{few_shot_feifei} entails the acquisition of a model endowed with the ability to classify novel classes with a limited number of labeled samples.
The mainstream few-shot image classification methods can be broadly divided into three categories: augmentation-based, optimization-based, and metric-based methods.
Augmentation-based attempts~\cite{aug_few_shot_1,aug_few_shot_2,aug_few_shot_3,aug_few_shot_4} usually design various augmentation strategies to expand the sample size to alleviate the data scarcity problem in few-shot settings.
Optimization-based techniques~\cite{MAML,MAML_1,MAML_2,MAML_3,MAML_4,MAML_5} design
optimization meta-learners to rapidly adapt model parameters to novel tasks with a few update steps, typical work like MAML~\cite{MAML}.
Metric-based methods~\cite{MatchNet,prototypical,RelationNet,few-shot-regular,ye2020few,TapNet} learn a common feature space and
apply a distance metric for few-shot matching.
%
Our method falls into the metric-based line but focuses on the more challenging few-shot action recognition task, which requires dealing with complex spatio-temporal structures.

\vspace{1mm}
\noindent \textbf{Caption generation.}
Traditional image/video captioning methods~\cite{caption_1,pan2017video,xu2015show,karpathy2015deep,anderson2018bottom,zhang2021rstnet,wang2018reconstruction,pei2019memory} usually employ an encoder-decoder architecture and construct the network by convolution, LSTM, or Transformer. 
However, these methods usually achieve unsatisfactory performance and poor transferability due to the limited amount of training data.
Recent studies~\cite{BLIP,zhang2021vinvl,li2020oscar,hu2022scaling,bao2021vlmo,wang2022git} have begun to explore web-scale image-text pairs for training and achieved remarkable progress.
%
Among these, BLIP~\cite{BLIP} proposes an effective end-to-end bootstrapping vision-language pretraining architecture, which removes the complicated object detector in feature extraction.
%
BLIP releases the pretrained models and also achieves superior performances in
effectiveness and efficiency across various
downstream tasks~\cite{chen2022litevl,santurkar2022caption,wang2022language,tiong2022plug}.
For simplicity and convenience, 
we employ BLIP for caption generation in CapFSAR.

\vspace{1mm}
\noindent \textbf{Multimodal foundation model.}
Recently, multimodal pretraining has been a popular paradigm to bridge vision and language
attributes and received tremendous success~\cite{li2021align,BLIP,CoCa,wang2022omnivl,zhai2022lit,CLIP,ALIGN,wang2023videocomposer,alayrac2022flamingo,yuan2022rlip}.
From the perspective of model structure, existing methods can be divided into encoder-only~\cite{CLIP,ALIGN,zhai2022lit} and encoder-decoder models~\cite{CoCa,wang2022git,BLIP}.
%
The former leverages cross-modal contrastive learning to align visual and text in the common embedding space, typical methods like CLIP~\cite{CLIP} and ALIGN~\cite{ALIGN}, which can
handle discriminative tasks, \eg, image-text retrieval and zero-shot classification.
The latter such as BLIP~\cite{BLIP} usually employs a generative text decoder to transform multimodal features into language, which is friendly to downstream captioning, visual question answers, \etc.
%
%
%

\begin{figure*}[ht]
  \centering
   \includegraphics[width=0.98\linewidth]{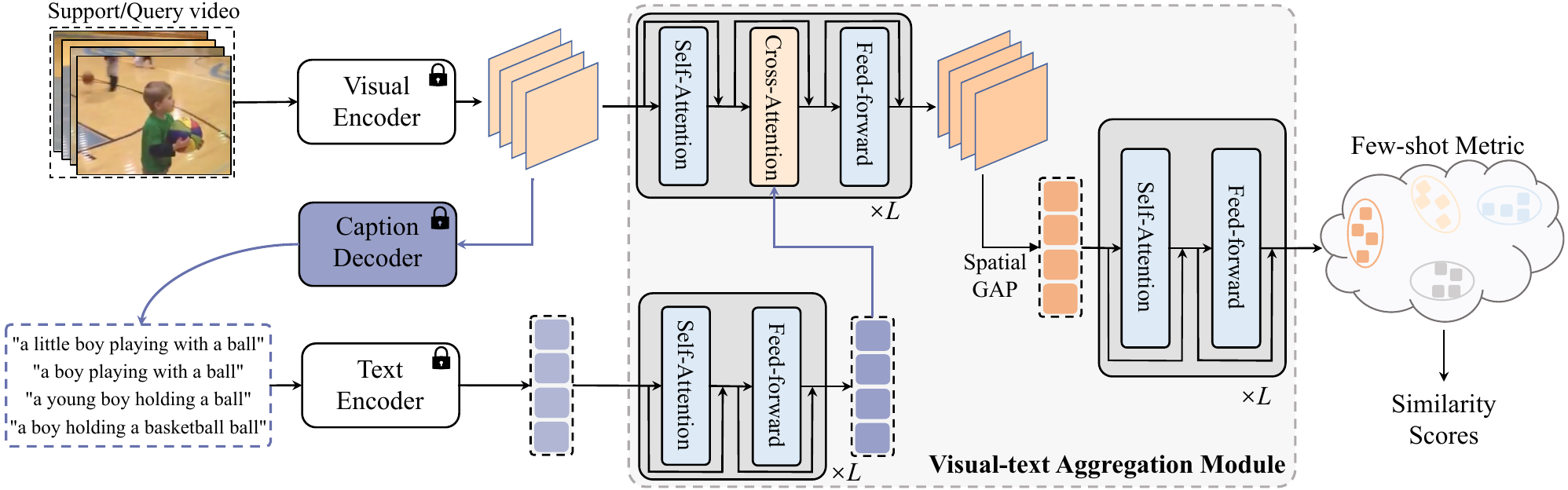}
\vspace{-1mm}
   \caption{
  Overall pipeline of CapFSAR. 
  Given an input video, a visual encoder is first applied to extract visual features.
  The caption decoder then generates captions on top of the image representations.
  %
  %
  Subsequently, a text encoder is leveraged to encode textual representations for these captions.
  Next, a visual-text aggregation module is further designed to fusion multimodal information. 
  Finally, the enhanced multimodal features are entered into a metric space to obtain the similarity scores of support-query pairs for few-shot classification.
  Note that the visual encoder, caption decoder, and text encoder are borrowed from the pretrained generative BLIP~\cite{BLIP} model.
  %
  %
  In order to preserve the original powerful knowledge, we freeze BLIP's parameters without updating them during training.
  %
  %
}
   \label{fig:framework}
   \vspace{-3mm}
\end{figure*}

\vspace{1mm}
\noindent \textbf{Few-shot action recognition.}
Most existing few-shot action recognition methods belong to the category of metric-based meta-learning due to its simplicity and effectiveness.
Previous work~\cite{CMN,CMN-J,OSS-metric,ARN-ECCV} mainly follows the idea of traditional few-shot image classification, which adopts a global-level matching strategy and ignores the temporal dynamic information in videos.
Subsequent methods~\cite{OTAM,ITANet,wang2023clip,wang2023cross,TRX,TA2N,HyRSM,STRM,wang2023hyrsmplus,MTFAN,huang2022compound,HCL,wang2023clip,nguyen2022inductive,wang2023molo} begin to explore mining the temporal relations of videos and designing various temporal alignment metrics.
Typically,
OTAM~\cite{OTAM} improves dynamic time warping~\cite{DTW} to explicitly utilize temporal path priors.
TRX~\cite{TRX} exhausts support-query frame matching combinations to deal with the problem of sub-action offset.
HyRSM~\cite{HyRSM} proposes a hybrid structure to aggregate temporal relations and designs a flexible temporal metric.
Huang \etal~\cite{huang2022compound} make use of object information and design two complementary temporal measures to perform few-shot matching. 
Even though these methods achieve remarkable results, they are still limited to using a single visual modality. While our CapFSAR attempts to explore multimodal information of foundation models and improves the classification accuracy in the low-shot regime.

\section{CapFSAR}


%
%
In this section, we detail the proposed framework called CapFSAR, which automatically generates captions for input videos and thus exploits multimodal information of foundation models for few-shot action recognition.
The overall architecture of CapFSAR is illustrated in Figure~\ref{fig:framework}. 
%
%

%
%

\subsection{Task Formulation}
In standard metric-based few-shot action recognition, there are two data sets, a meta-train set for training the few-shot model and a meta-test set for testing, whose label spaces are disjoint.
Generally, in order to simulate the few-shot test environment, training and testing of this task are composed of many $N$-way $K$-shot tasks/episodes~\cite{OTAM,TRX,HyRSM}.
In each $N$-way $K$-shot task, there is a support set $S=\{s_{1},s_{2},...,s_{N \times K}\}$ containing randomly sampled $N$ action classes and each class consists of
$K$ labeled videos, and a query set $Q$ comprises videos to be tested.
The purpose is to classify the query videos in $Q$ into one of the $N$ classes according to the labeled support set.
%

\subsection{Caption Generation via BLIP}

Since the input videos only contain visual information and no textual description is involved, 
 we need to generate the textual descriptions automatically.
%
We accomplish this goal by leveraging publicly available state-of-the-art BLIP~\cite{BLIP}, a large-scale vision-language pretrained model.
The core components of the pretrained BLIP model used in CapFSAR mainly include three parts: a visual encoder, a caption decoder, and a text encoder.
%
%

%
\textbf{Visual encoder.} 
The goal of this encoder is to extract a compact visual representation for the video.
Given an input video, following the previous methods~\cite{OTAM,TRX}, we first perform a sparse sampling strategy~\cite{TSN} to save computation and extract a temporal sequence of video frames $v=\{v^1, v^2, ..., v^T\}$, where $T$ is the temporal length.
%
%
Then, we apply the visual encoder of BLIP to generate visual features $f_{v}=\{f^{1}_{v},f^{2}_{v},...,f^{T}_{v} \} \in \mathbb{R}^{T\times C\times H\times W}$, where $C$, $H$, $W$ represent  channels, height, and width respectively.

\textbf{Caption decoder.}
This module takes the visual features as input and predicts the corresponding language descriptions.
Specifically, a \texttt{[Start]} token is first used to signal the beginning of the predicted sequence. 
Then, the sequence interacts with visual features through cross-attention and  recursively predicts the next word.
Finally, a \texttt{[End]} token is leveraged to denote the end of the prediction.
To generate captions, 
we input the visual feature $f_{v}$ into the caption decoder and express the output descriptions as $\textrm{Cap}_{v}$:
\begin{equation}
\text{Cap}_{v} = \mathrm{Decoder}(f_{v})
\end{equation}
where $\textrm{Cap}_{v}=\{\textrm{Cap}^{1}_{v}, \textrm{Cap}^{1}_{v}, ...,\textrm{Cap}^{T}_{v}\}$ and $\textrm{Cap}^{i}_{v}$ represents the generated caption for frame $v^{i}$.

\textbf{Text encoder.}
The synthetic captions can be leveraged as an interface to extract the rich contextual knowledge in the large language model.
They can be viewed as an additional perspective to supplement visual information.
For convenience, we adopt the off-the-shelf text encoder of BLIP as a language model to encode textual representations.
The model architecture of the text encoder is a Transformer-based BERT~\cite{BERT}.
To summarize the input sentence, a \texttt{[CLS]} token is appended to the beginning of the caption, and the resulting token feature can be used to encode the overall caption.
The formula is expressed as:
\begin{equation}
\textrm{\textbf{t}}_{v} = \mathrm{Encoder}_{text}(\textrm{Cap}_{v})
\end{equation}
%
where $\textrm{\textbf{t}}_{v}\in \mathbb{R}^{T\times C}$ denotes the output text representations.

\subsection{Visual-text Aggregation Module}
How to aggregate visual and textual information is essential for the final performance since they contain abundant spatio-temporal dependencies.
We design a visual-text aggregation module to perform cross-modal interaction between the visual and text features and further enhance the video representation.
The key idea of this module is to fully mine the temporal information of text and video and incorporate textual features to spatially modulate visual features, as captions often contain details of interest~\cite{yuan2022rlip}, such as subjects, objects, and human-object interactions.

Concretely, 
we first apply text Transformers composed of self-attention and feed-forward blocks to text features for temporal modeling, which aims to boost the context awareness of text representations.
Then, the obtained text features are leveraged to spatially modulate visual features and fuse multimodal information through the cross-attention operator.
Finally, the resulting features are spatially collapsed through spatial global average-pooling (Spatial GAP) and fed into temporal transformers to aggregate temporal correlations. %
For simplicity, we set all Transformer layers to the same number of $L$.
We formulate the whole process as:
\begin{equation}
{\widetilde{f}}_{v} = \mathcal{VT}(f_{v}, \textrm{\textbf{t}}_{v})
\end{equation}
where $\mathcal{VT}$ represents the visual-text aggregation module and ${\widetilde{f}}_{v} \in \mathbb{R}^{T\times C}$ is the output discriminative multimodal features, which have integrated visual and textual cues and contain diverse spatio-temporal relations within the video.
%
%
%
%

\subsection{Few-shot Metric}
After obtaining the multimodal features of the support and query videos in a few-shot task, like prior works~\cite{OTAM,TRX,HyRSM}, we apply a temporal metric such as OTAM~\cite{OTAM} to generate the final support-query matching results:
%
%
\begin{equation}
{{D}}_{S, q} = \mathcal{M}(\widetilde{f}_{S}, \widetilde{f}_{q})
\end{equation}
where $\widetilde{f}_{S}$ represents the support features, $\widetilde{f}_{q}$ is the query features,
and
$\mathcal{M}$ denotes the temporal metric module to calculate the support-query similarity scores ${{D}}_{S, q}$ along the temporal dimension.
%
Since our CapFSAR is a plug-and-play algorithm, we can directly utilize the metrics and training objectives of existing methods to validate our approach, including OTAM~\cite{OTAM}, TRX~\cite{TRX}, and HyRSM~\cite{HyRSM}.
%
%

%

\section{Experiments}

To comprehensively evaluate and validate the effectiveness of our approach, we perform extensive comparative experiments with state-of-the-art methods and detailed ablation studies on multiple publicly available datasets.

\subsection{Experimental Setup}

\noindent \textbf{Datasets.} 
%
We conduct evaluation experiments on five commonly used public benchmarks and follow the common practice~\cite{OTAM,TRX,HyRSM} to pre-process the datasets for a fair comparison.
%
For Kinetics~\cite{Kinetics} and SSv2-Full~\cite{SSv2}, we adopt the splits from \cite{OTAM,HyRSM}, where 64/24 classes are used for training/testing.
We also utilize the SSv2-Small dataset proposed in ~\cite{CMN}, which is smaller than SSv2-Full and contains 100 videos per class.
For HMDB51 and UCF101, we follow the settings of ~\cite{ARN-ECCV,HyRSM} to have a fair comparison.
%

\noindent \textbf{Evaluation protocol.} 
Following the existing standard few-shot protocol~\cite{OTAM,HyRSM}, we evaluate CapFSAR under 5-way 1-shot and 5-way 5-shot setups.
We randomly select 10,000 episodes from the test set and report the average accuracy.

\noindent \textbf{Implementation details.} 
In the experiments, we utilize the openly available BLIP$_{\textrm{ViT-B}}$~\cite{BLIP,ViT} model pretrained on 129M image-text pairs by default and apply the beam search~\cite{gulcehre2015using} sampling strategy to generate captions for each video.
The BLIP model is frozen during the training phase and will not update parameters, so we pre-extract frame captions offline to save training time.
For a fair comparison with previous methods~\cite{OTAM,HyRSM,HCL}, we also conduct experiments with the widely used ResNet-50~\cite{Resnet} backbone pretrained on ImageNet~\cite{imagenet}. 
%
We randomly uniformly sample $T=8$ frames for frame extraction and crop a $224 \times 224$ region as input. 
CapFSAR is optimized via Adam~\cite{adam} and trained by the PyTorch library.
For 5-shot evaluation, we follow the original feature fusion principles of ~\cite{OTAM,TRX,HyRSM} to obtain the final class prototypes for classification.


%
%
\begin{table*}[t]
\centering
\small
\tablestyle{5.5pt}{1.06}
\begin{tabular}
{l|c|c|cc|cc|cc|cc|cc}
\hline
\hspace{-0.5mm}  \multirow{2}{*}{Method} \hspace{2mm} & \multicolumn{1}{c|}{\multirow{2}{*}{Venue}} &
\multicolumn{1}{c|}{\multirow{2}{*}{Backbone}}
&\multicolumn{2}{c|}{{Kinetics}} 
&\multicolumn{2}{c|}{{SSv2-Full}} &\multicolumn{2}{c|}{{UCF101}}  &\multicolumn{2}{c|}{{SSv2-Small}}  & \multicolumn{2}{c}{{HMDB51}}  \\
& \multicolumn{1}{c|}{} & \multicolumn{1}{c|}{} 
& \multicolumn{1}{l}{1-shot}  & 5-shot & \multicolumn{1}{l}{1-shot}   & 5-shot 
& \multicolumn{1}{l}{1-shot}   & 5-shot
& \multicolumn{1}{l}{1-shot}   & 5-shot
& \multicolumn{1}{l}{1-shot}   & 5-shot  \\ 
\shline
MatchingNet~\cite{MatchNet}    & NIPS'16 & INet-RN50
& 53.3   & 74.6
& - & -
& -   & -   
& 31.3    & 45.5   
& -  & -    \\
MAML~\cite{MAML}    & ICML'17& INet-RN50
& 54.2   & 75.3 
& -   & - 
& -   & -   
& 30.9    & 41.9   
& -   & -    \\
Plain CMN~\cite{CMN}    & ECCV'18& INet-RN50
& 57.3   & 76.0
& -   & - 
& -   & -   
& 33.4    & 46.5   
& -   & -    \\
CMN++~\cite{CMN}    & ECCV'18& INet-RN50
& 65.4  & 78.8 
& 34.4   & 43.8 
& -   & -   
& -    & -   
& -   & -    \\
TRN++~\cite{TRN-ECCV}    & ECCV'18 & INet-RN50
& 68.4   & 82.0 
& 38.6   & 48.9 
& -   & -   
& -    & -   
& -   & -    \\
TARN~\cite{TARN} & BMVC’19 & C3D
& 64.8   & 78.5 
& -   & - 
& -   & -   
& -    & -   
& -   & -    \\
CMN-J~\cite{CMN-J}    & TPAMI'20& INet-RN50
& 60.5   & 78.9
& -   & - 
& -    & -   
& 36.2    & 48.8   
& -   & -    \\
ARN~\cite{ARN-ECCV}    & ECCV'20 & C3D
& 63.7   & 82.4
& -   & - 
& 66.3    & 83.1   
& -     & -   
& 45.5   & 60.6    \\
OTAM~\cite{OTAM}    & CVPR'20& INet-RN50
& 73.0    & 85.8 
& 42.8   & 52.3 
& 79.9    & 88.9   
& 36.4    & 48.0   
& 54.5    & 68.0    \\
ITANet~\cite{ITANet}    & IJCAI'21& INet-RN50
& 73.6   & 84.3 
& 49.2   & 62.3 
& -    & -   
& 39.8    & 53.7  
& -   & -    \\
TRX~\cite{TRX}    & CVPR'21& INet-RN50
& 63.6   & 85.9 
& 42.0   & 64.6 
& 78.2   & {96.1}   
& 36.0   & \hspace{0.6mm} 56.7$^{*}$  
& 53.1   & 75.6    \\
TA$^{2}$N~\cite{TA2N}    & AAAI'22& INet-RN50
& 72.8   & 85.8 
& 47.6   & 61.0 
& 81.9   & 95.1   
& -   & -   
& 59.7    & 73.9    \\
MTFAN~\cite{MTFAN}    & CVPR'22& INet-RN50
& 74.6   & 87.4
& 45.7   & 60.4
& 84.8   & 95.1   
& -    & -  
& 59.0   & 74.6    \\
STRM~\cite{STRM}    & CVPR'22& INet-RN50
& 62.9   & 86.7 
& 43.1   & 68.1 
& 80.5  & \underline{96.9}
& 37.1 & 55.3
& 52.3  & {77.3}   \\
HyRSM~\cite{HyRSM}    &  CVPR'22& INet-RN50
& 73.7   & 86.1 
& \underline{54.3}   & \underline{69.0} 
& 83.9    & 94.7   
& \underline{40.6}    & 56.1  
& {60.3}    & 76.0    \\
%
Nguyen~\etal~\cite{nguyen2022inductive}    & ECCV'22& INet-RN50
& 74.3   & 87.4 
& 43.8   & 61.1 
& 84.9    & 95.9   
& -   & -  
& 59.6   & {76.9}    \\
Huang~\etal~\cite{huang2022compound}    & ECCV'22& INet-RN50
& 73.3   & 86.4 
& 49.3   & 66.7 
& 71.4    & 91.0   
& 38.9    & \textbf{{61.6}}  
& 60.1   & {77.0}    \\ 
HCL~\cite{HCL}    & ECCV'22& INet-RN50
& 73.7   & 85.8 
& 47.3   & 64.9
& 82.5    & 93.9   
& 38.7    & 55.4  
& 59.1    & 76.3    \\ 
%
MoLo~\cite{wang2023molo}    & CVPR'23 & INet-RN50
& 73.8   & 85.1 
& 55.0   & \textbf{69.6}
& 85.4    & 95.1  
& \textbf{41.9}    & 56.2  
& 59.8    & 76.1   \\ 
%
\rowcolor{Gray}
\textbf{CapFSAR (OTAM)}    & - & INet-RN50
& \underline{79.2}   & {88.8} 
& 48.5   & 65.0 
& \underline{89.0}   & 96.2
& 40.0    & 55.1
& 59.9    & 73.7    \\
\rowcolor{Gray}
\textbf{CapFSAR (TRX)}    & - & INet-RN50
& 71.8   & \textbf{89.1} 
& 47.5   & 65.2 
& 88.8   & \textbf{97.0}
& 38.4    & \underline{57.0}
& \underline{63.0}    & \textbf{79.1}    \\
\rowcolor{Gray}
\textbf{CapFSAR (HyRSM)}    & - & INet-RN50
& \textbf{79.3}   & \underline{89.0} 
& \textbf{55.1}   & \textbf{69.6} 
& \textbf{89.2}   & 95.6
& {41.1}   & 56.7
& \textbf{64.1}   & \underline{77.6}    \\
\shline
BLIP-Freeze$_{\textrm{visual}}$~\cite{BLIP}    & ICML'22 & BLIP$_{\textrm{ViT-B}}$
& 74.8   & 87.5 
& 31.0   & 44.6 
& 88.9  & 95.3   
& 31.2  & 40.3
& 56.2  & 72.8     \\
BLIP-Freeze$_{\textrm{text}}$~\cite{BLIP}    & ICML'22 & BLIP$_{\textrm{ViT-B}}$
& 72.9   & 86.5 
& 29.8   & 41.1 
& 86.4  & 95.1   
& 28.7  & 39.5
& 52.4  & 67.2     \\
OTAM$^\dagger$    & CVPR'20 & BLIP$_{\textrm{ViT-B}}$
& 82.4 & 91.1
& 50.2   & 65.3
&  91.4  & 96.5   
& 45.5  & 58.7  
& 63.9  & 76.5     \\
TRX$^\dagger$    & CVPR'21 & BLIP$_{\textrm{ViT-B}}$
& 76.6   & 90.8 
& 45.1  & 68.5 
& 90.9   & 97.4
& 40.6    & 61.0
& 58.9    & 79.9    \\
HyRSM$^\dagger$    & CVPR'22 & BLIP$_{\textrm{ViT-B}}$
& 82.4   & 91.8 
& \underline{52.1}   & 69.5 
& 91.6   & 96.9
& 45.5    & 60.7
& \underline{69.8}    & \underline{80.6}    \\
\rowcolor{Gray}
\textbf{CapFSAR (OTAM)}     & -& BLIP$_{\textrm{ViT-B}}$
& \textbf{84.9}   & \textbf{93.1}
& 51.9   & 68.2
& \textbf{93.3}  & \underline{97.8}  
& \textbf{45.9}   & 59.9
& 65.2  & 78.6    \\
\rowcolor{Gray}
\textbf{CapFSAR (TRX)}    & - & BLIP$_{\textrm{ViT-B}}$
& 78.1   & 91.2 
& 47.2   & \underline{69.7} 
& 92.1   & \textbf{97.9}
& 42.3    & \textbf{61.7}
& 62.3    & 80.4     \\
\rowcolor{Gray}
\textbf{CapFSAR (HyRSM)}    & - & BLIP$_{\textrm{ViT-B}}$
& \underline{83.5}   & \underline{92.2} 
& \textbf{54.0}   & \textbf{70.1} 
& \underline{93.1}   & 97.7
& \underline{45.8}    & \underline{61.1}
& \textbf{70.3}    & \textbf{81.3}    \\
\hline
\end{tabular}
\vspace{+0.5mm}
\caption{Comparison to existing state-of-the-art few-shot action recognition techniques on the Kinetics, SSv2-Full, UCF101, SSv2-Small,  and HMDB51 datasets. 
%
The experiments are conducted under the 5-way 1-shot and 5-way 1-shot settings.
%
``-" means the result is not available in previously published works.
%
``${*}$"  represents the results of our implementation.
The best results are in bold, and the second-best ones are underlined. 
``INet-RN50" denotes ResNet-50 pretrained on the ImageNet~\cite{imagenet} dataset.
%
``$\dagger$" stands for visual-text aggregation module without text branch is also implemented for a fair comparison (\ie, baseline methods displayed in Figure~\ref{fig:baseline}(a)).
%
}
\label{tab:compare_SOTA_1}
\vspace{-4mm}
\end{table*}

\begin{figure}[t]
  \centering
   \includegraphics[width=0.98\linewidth]{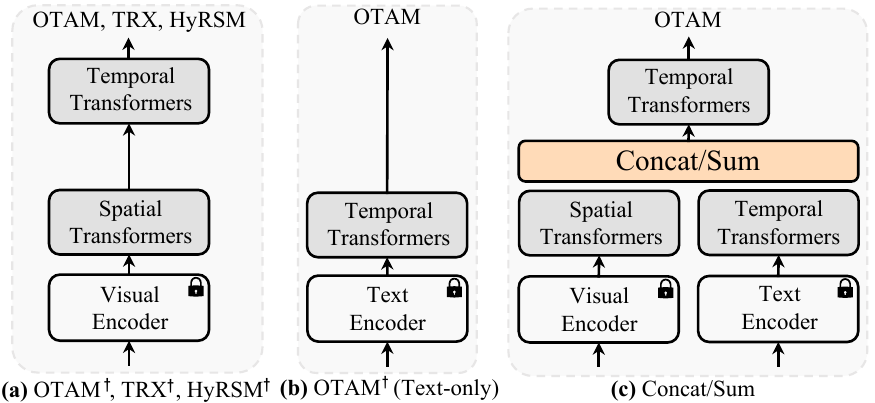}
\vspace{-0mm}
   \caption{
  Schematic diagram of baseline methods and variants.
  %
}
   \label{fig:baseline}
   \vspace{-3mm}
\end{figure}

\noindent \textbf{Baselines.} 
To demonstrate the superiority of the proposed method,
we compare CapFSAR with previous state-of-the-art methods, including OTAM~\cite{OTAM}, TRX~\cite{TRX}, MTFAN~\cite{MTFAN}, STRM~\cite{STRM} HyRSM~\cite{HyRSM}, HCL~\cite{HCL}, \etc.
Following the literature~\cite{OTAM,TRX,STRM,HyRSM,HCL}, we adopt ImageNet pretrained ResNet-50 as the visual encoder for fair comparison and leverage BLIP for additional caption generation.
Moreover, since no previous few-shot works adopt the BLIP model for few-shot classification, we additionally conduct comparative experiments with the following two types of baselines: 
\textit{i}) We introduce a BLIP-Freeze method, which implements BLIP's frozen visual encoder (BLIP-Freeze$_{\textrm{visual}}$) or text encoder (BLIP-Freeze$_{\textrm{text}}$) as the backbone to encode features and applies OTAM for few-shot matching;
\textit{ii}) To further verify the effectiveness of our CapFSAR,
as depicted in Figure~\ref{fig:baseline}(a),
we also construct three stronger baselines, namely OTAM$^\dagger$, TRX$^\dagger$ and HyRSM$^\dagger$, which use the same BLIP's visual encoder as CapFSAR and utilize the visual-text aggregation module without text branch (\ie, single visual modality) by default for a fair comparison.

\subsection{Comparison with State-of-the-Arts}
As mentioned earlier, the proposed CapFSAR is a plug-and-play framework. We insert CapFSAR to three existing representative works whose source code is available, \ie, OTAM~\cite{OTAM}, TRX~\cite{TRX} and HyRSM~\cite{HyRSM}, and conduct comparative experiments with current state-of-the-art methods.
Table~\ref{tab:compare_SOTA_1} summarizes the detailed
comparison results on five common benchmarks under the 5-way 1-shot and 5-way 5-shot settings. 
From the results, we can observe that when using the same ImageNet pretrained ResNet-50 backbone, CapFSAR achieves significant improvements over the three baselines in terms of all metrics and outperforms previous state-of-the-art methods by a convincing margin.
Specifically, CapFSAR based on HyRSM reaches 79.3\% 1-shot accuracy, which boosts the original HyRSM by 5.6\% and displays superior performance over previous state-of-the-art MTFAN~\cite{MTFAN}.
Note that under the 5-shot SSv2-Small setting, our CapFSAR lags behind  Huang~\etal~\cite{huang2022compound}, possibly because Huang~\etal utilize multiple measurements for ensemble.
Since no publicly released code is available, we can't plug CapFSAR into ~\cite{huang2022compound}, but we believe our framework is generic.
%
%
In addition, compared with HyRSM, the proposed CapFSAR based on HyRSM has a relatively slight improvement on the SSv2-Full and SSv2-Small datasets because HyRSM specializes in temporal modeling and performs well enough on these two datasets.
To further validate the superiority of CapFSAR, we extend existing methods with the BLIP$_{\textrm{ViT-B}}$ model, \ie, OTAM$^\dagger$, TRX$^\dagger$, and HyRSM$^\dagger$, and compare them with CapFSAR.
%
Our CapFSAR still displays the best performance among all comparison methods.
For example, based on OTAM, CapFSAR achieves 78.6\% 1-shot performance on HMDB51, which brings 2.1\% improvement over OTAM$^\dagger$, indicating the effectiveness of incorporating  textual information to assist few-shot action recognition.
Moreover, by comparing CapFSAR with the other three counterparts, we notice that the proposed CapFSAR brings convincing gains across all datasets, illustrating the applicability of our pipeline.

%
%
\begin{table}[t]
\centering
\small
\tablestyle{6pt}{1.1}
\begin{tabular}
{l|c|cc|cc}
\hline
			
\hspace{-0.5mm}  \multirow{2}{*}{Method} \hspace{2mm} &   \multirow{2}{*}{Modality}
&\multicolumn{2}{c|}{{Kinetics}} 
&\multicolumn{2}{c}{{SSv2-Full}}  \\
& & \multicolumn{1}{l}{1-shot}  & 5-shot & \multicolumn{1}{l}{1-shot}   & 5-shot 
 \\ 
\shline
BLIP-Freeze & Visual-only
& 74.8   & 87.5
& 31.0 & 44.6 \\
OTAM$^\dagger$ & Visual-only
& 82.4   & 91.1
& 50.2 & 65.3 \\
BLIP-Freeze & Text-only
& 72.9   & 86.5
& 29.8 & 41.1 \\
OTAM$^\dagger$ & Text-only
& 78.3   & 88.3
& 36.4 & 48.2 \\
\rowcolor{Gray}
\textbf{CapFSAR} & Multimodal
& \textbf{84.9}   & \textbf{93.1}
& \textbf{51.9} & \textbf{68.2}
 \\
\hline
\end{tabular}
\vspace{+1mm}
\caption{
Ablation study on the Kinetics and SSv2-Full datasets regarding 5-way 1-shot and 5-way 5-shot accuracy.
%
``Text-only BLIP-Freeze" indicates that text features output by text encoder are directly classified using OTAM without involving learnable modules.
OTAM$^\dagger$ means the baseline method in Figure~\ref{fig:baseline}.
}
\label{tab:ablation}
\vspace{-1mm}
\end{table}


%
%
\begin{table}[t]
\centering
\small
\tablestyle{6pt}{1.1}
\setlength{\tabcolsep}{1.5mm}{
\begin{tabular}
{l|cc|cc}
\hline
			
\hspace{-0.5mm}  \multirow{2}{*}{Aggregation manner} \hspace{2mm} &   \multicolumn{2}{c|}{{Kinetics}} 
&\multicolumn{2}{c}{{SSv2-Full}}  \\
 & \multicolumn{1}{l}{1-shot}  & 5-shot & \multicolumn{1}{l}{1-shot}   & 5-shot 
 \\ 
\shline
Concat
& 84.6   & 92.9
& 51.2 & 68.0 \\
Sum
& 84.5   & 92.4
& 51.2 & 67.3 \\
\textbf{Cross-Attention (CapFSAR)}
& \textbf{84.9}   & \textbf{93.1}
& \textbf{51.9} & \textbf{68.2} \\
\hline
\end{tabular}}
\vspace{+1mm}
\caption{Ablation study on different aggregation manners.
}
\label{tab:Aggregation_manner}
\vspace{-3mm}
\end{table}

\subsection{Ablation Study}
%

We conduct comprehensive ablation studies on multiple
benchmarks to investigate the capability of the proposed CapFSAR and analyze the role of each component.
Unless otherwise specified, CapFSAR based on OTAM~\cite{OTAM} with BLIP$_{\textrm{ViT-B}}$ is adopted as the default setting for ablation.

\vspace{1mm}
\noindent \textbf{Importance of multimodal fusion.}
To investigate the role of visual-text aggregation, we conduct experiments to ablate each modal branch in Table~\ref{tab:ablation}.
``BLIP-Freeze" means that the visual features (visual-only) or text features (text-only) output by the respective encoders are directly input to the OTAM~\cite{OTAM} for classification.
The visual-only OTAM$^\dagger$ and text-only OTAM$^\dagger$ correspond to the methods in Figure~\ref{fig:baseline}(a) and (b), respectively.
From the results, we can find that visual-only methods generally  outperform the text-only counterparts, which we attribute to visual modality containing more local details while the text is a global overview. 
In addition, by fusing multimodal information, CapFSAR achieves the highest performance, \eg, 68.2\% 5-shot result on the SSv2-Full dataset, which is 2.9\% ahead of the visual-only OTAM$^\dagger$.
This fully reveals that the text descriptions automatically generated by the caption decoder can provide an augmented view for the input video and complement the visual features, which is consistent with our motivation.
\begin{table}[t]
\centering
\small
\tablestyle{6pt}{1.1}
\setlength{\tabcolsep}{1.5mm}{
\begin{tabular}
{l|cc|cc}
\hline
			
\hspace{-0.5mm}  \multirow{2}{*}{Transformer layers $L$} \hspace{2mm} &   \multicolumn{2}{c|}{{Kinetics}} 
&\multicolumn{2}{c}{{SSv2-Full}}  \\
 & \multicolumn{1}{l}{1-shot}  & 5-shot & \multicolumn{1}{l}{1-shot}   & 5-shot 
 \\ 
\shline
$L=1$ (Default)
& \textbf{84.9}   & \textbf{93.1}
& \textbf{51.9} & 68.2 \\
$L=2$
& 83.3   & 92.0
& 50.7 & 68.8 \\
$L=3$
& 82.2   & 91.6
& 51.5 & \textbf{69.5} \\
$L=4$
& 82.1   & 91.1
& 51.3 & 67.6 \\
\hline
\end{tabular}}
\vspace{+1mm}
\caption{Ablation study on the effect of Transformer layers $L$.
}
\label{tab:transformer_number}
\vspace{-2mm}
\end{table}
%
%
%
\begin{table}[t]
\centering
\small
\tablestyle{6pt}{1.1}
\begin{tabular}
{l|cc|cc}
\hline
			
\hspace{-0.5mm}  \multirow{2}{*}{Setting} \hspace{2mm} &   \multicolumn{2}{c|}{{Kinetics}} 
&\multicolumn{2}{c}{{SSv2-Full}}  \\
 & \multicolumn{1}{l}{1-shot}  & 5-shot & \multicolumn{1}{l}{1-shot}   & 5-shot 
 \\ 
\shline
w/o text temporal Transformer
&  84.3  & 92.9 
&  51.1 &  67.6 \\
\textbf{CapFSAR}
&  \textbf{84.9}  & \textbf{93.1} 
&  \textbf{51.9} &  \textbf{68.2} \\
\hline
\end{tabular}
\vspace{-1mm}
\caption{Experiments on the impact of text temporal Transformer.
}
\label{tab:text_transformer}
\vspace{-3mm}
\end{table}

\vspace{1mm}
\noindent \textbf{Effect of different aggregation manners.}
In the visual-text aggregation module, we propose to spatially fuse textual and visual features through the cross-attention operator.
Table~\ref{tab:Aggregation_manner} reports the comparisons of different aggregation manners.
%
As depicted in Figure~\ref{fig:baseline}(c),
``Concat$/$Sum" means that the visual features output by spatial Transformers and the text features output by temporal Transformers are directly concatenated$/$summed and then fed into Temporal Transformers for multimodal fusion.
Among them, the cross-attention variant achieves the consistently best results suggesting the effectiveness of our module design.

\vspace{1mm}
\noindent \textbf{Influence of the Transformer layers $L$.}
In order to explore the impact of different $L$ on performance, we conduct ablation experiments on the Kinetics and SSv2-Full datasets in Table~\ref{tab:transformer_number}.
On the Kinetics dataset, the best results are obtained on 1-shot and 5-shot when $L=1$, and overfitting starts to occur as $L$ increases due to the fact that this dataset is appearance-biased and relatively easy to identify~\cite{TSM,HyRSM}.
%
On the complex motion-biased SSv2-Full dataset, the best 1-shot performance is achieved when $L=1$, and the best 5-shot result is reached when $L=3$. 
To balance accuracy and efficiency, we choose $L=1$ as our default setting.

%
%

\vspace{1mm}
\noindent \textbf{Effect of the text temporal Transformer.}
In CapFSAR, the text Transformer is adopted to extract temporal context for the input captions.
%
We explore the effect of this component in Table~\ref{tab:text_transformer}.
Compared with CapFSAR, the method without the text transformer leads to inferior performance, such as the 1-shot SSv2 result drops from 51.9\% to 51.1\%. 
This illustrates the importance of using the Transformer to improve temporal awareness of text representations.

\begin{figure}[t]
  \centering
   \includegraphics[width=0.98\linewidth]{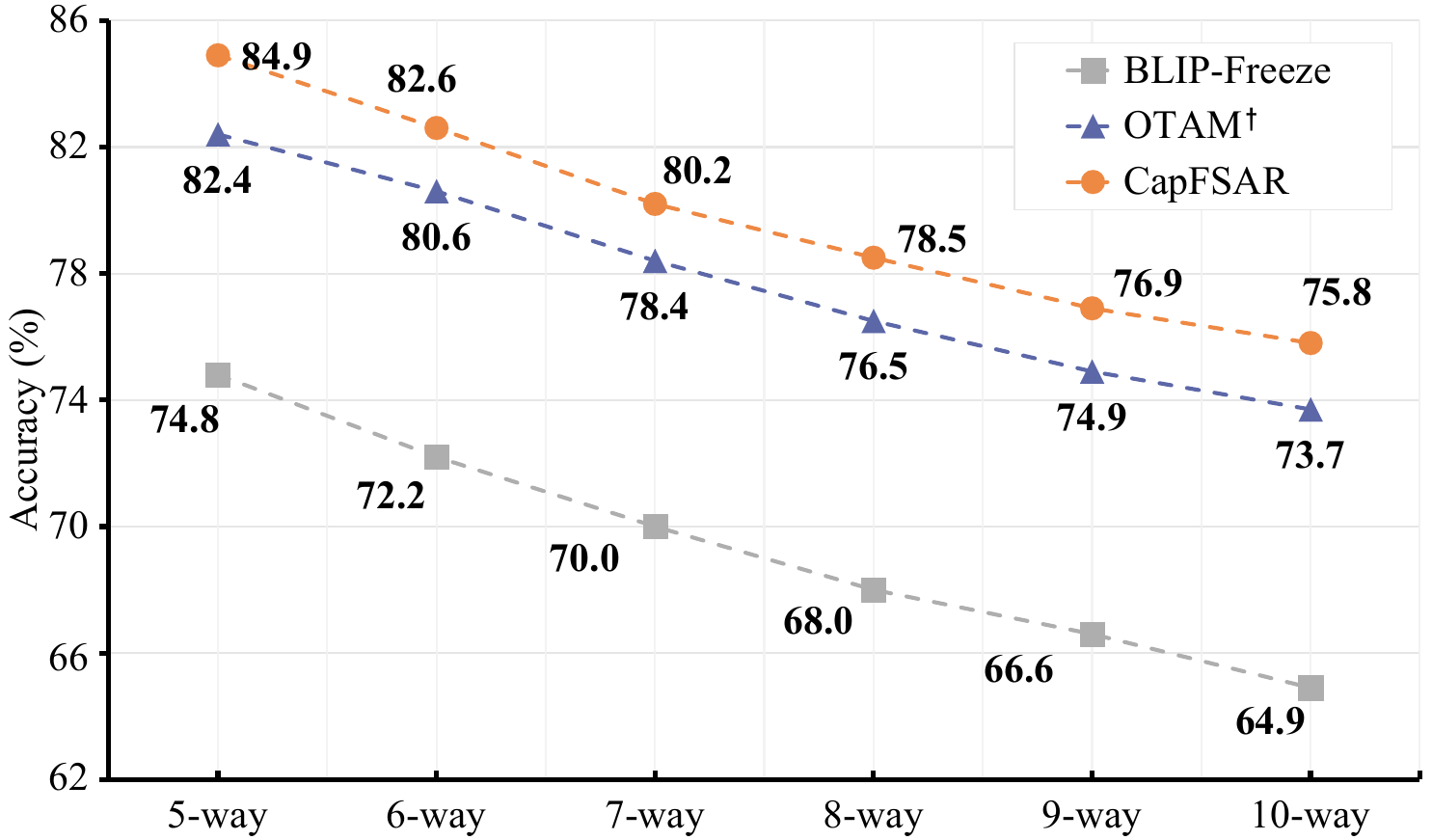}
\vspace{-1mm}
   \caption{
  $N$-way 1-shot experiment on the Kinetics dataset.
}
   \label{fig:way_ablation}
\end{figure}

\vspace{1mm}
\noindent \textbf{$N$-way classification.}
The previous experiments are all performed on the 5-way setting. 
In order to further analyze the impact of different ways on performance, we conduct $N$-way 1-shot ablation. 
As presented in Figure~\ref{fig:way_ablation},
we notice that compared to the baseline methods, our CapFSAR achieves consistent superior performance under various settings, illustrating the scalability of the proposed method.

\begin{figure}[t]
  \centering
   \includegraphics[width=0.98\linewidth]{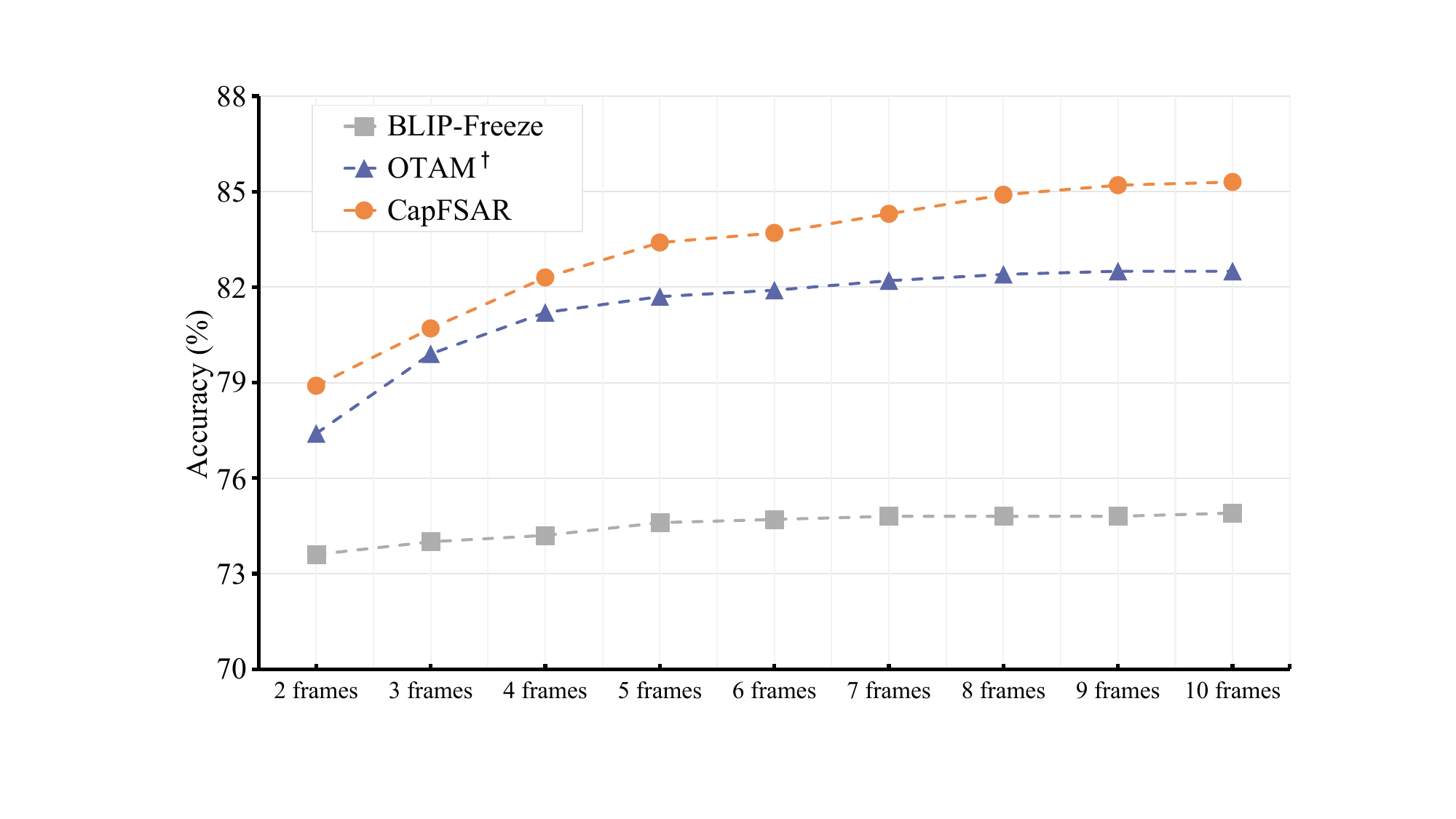}
\vspace{-1mm}
   \caption{
  Ablation experiment with varying the number of input frames under the 5-way 1-shot setting on the Kinetics dataset.
}
   \label{fig:frame_ablation}
   \vspace{-2mm}
\end{figure}

\vspace{1mm}
\noindent \textbf{Varying the number of input frames.}
We thoroughly investigate the impact of sampling different input frame numbers on the few-shot performance in Figure~\ref{fig:frame_ablation}.
%
We have the following two findings: i) As the number of input video frames increases, the performance gradually improves and eventually tends to be saturated due to visual information redundancy;
ii) Our CapFSAR consistently outperforms the baselines, and the performance improvement is more remarkable when the number of input frames is large. 
%
We attribute this to the fact that the increase in caption information can significantly supplement the visual representations, yielding more discriminative multimodal features.

\vspace{1mm}
\noindent \textbf{Effect of diverse captions.}
In our default setting, we synthesize captions by beam search~\cite{gulcehre2015using}, a deterministic decoding technique
that produces only one description with the highest probability.
In Table~\ref{tab:beam_manner}, we additionally employ the stochastic nucleus sampling~\cite{holtzmancurious}  to synthetic more diverse text descriptions and decode five captions per frame for comparison.
We can find that the nucleus sampling strategy generally obtains higher performance than beam search algorithm.
For instance, nucleus sampling reaches 69.1\% 5-shot result on SSv2-Full, surpassing beam search by 0.9\%,
which illustrates that more diverse captions can provide more additional algorithm information and thus boost the classification results. 
The above observation is also in line with our intuition that the generated textual descriptions can help to produce more comprehensive multimodal features.
%

%
%
\begin{table}[t]
\centering
\small
\tablestyle{6pt}{1.1}
\begin{tabular}
{l|cc|cc}
\hline
			
\hspace{-0.5mm}  \multirow{2}{*}{Generation strategy} \hspace{2mm} &   \multicolumn{2}{c|}{{Kinetics}} 
&\multicolumn{2}{c}{{SSv2-Full}}  \\
 & \multicolumn{1}{l}{1-shot}  & 5-shot & \multicolumn{1}{l}{1-shot}   & 5-shot 
 \\ 
\shline
Beam Search (Default)
& 84.9   & 93.1
& 51.9 & 68.2 \\
Nucleus Sampling
& \textbf{85.2}   & \textbf{93.2}
& \textbf{52.1}   & \textbf{69.1 } \\
\hline
\end{tabular}
\vspace{+1mm}
\caption{Ablation study on different caption generation strategies.
}
\label{tab:beam_manner}
\vspace{-1mm}
\end{table}

%
%
\begin{table}[t]
\centering
\small
\tablestyle{6pt}{1.1}
\setlength{\tabcolsep}{1.5mm}{
\begin{tabular}
{l|c|cc|cc}
\hline
			
\hspace{-0.5mm}  \multirow{2}{*}{Method} \hspace{2mm} &    {Pretrained}  &  \multicolumn{2}{c|}{{Kinetics}} 
&\multicolumn{2}{c}{{SSv2-Full}}  \\
 &  data & \multicolumn{1}{l}{1-shot}  & 5-shot & \multicolumn{1}{l}{1-shot}   & 5-shot 
 \\ 
\shline
BLIP$_{\textrm{ViT-B}}$ & 14M
&  84.1  &  92.6
&  51.5  &  68.1 \\
BLIP$_{\textrm{ViT-B}}$ (Default)  & 129M
&   84.9  &  {93.1}
&   51.9  &  68.2 \\
BLIP$_{\textrm{ViT-L}}$  & 129M
&  \textbf{85.2}   &  \textbf{93.2}
&  \textbf{52.0}   &  \textbf{68.3} \\
\hline
\end{tabular}}
\vspace{+1mm}
\caption{Ablation study on different caption generation methods.
}
\label{tab:caption_method}
\vspace{-1mm}
\end{table}

%
%
\begin{table}[t]
\centering
\small
\tablestyle{6pt}{1.1}
\begin{tabular}
{l|cc|cc}
\hline
			
\hspace{-0.5mm}  \multirow{2}{*}{Text encoder} \hspace{2mm} &   \multicolumn{2}{c|}{{Kinetics}} 
&\multicolumn{2}{c}{{SSv2-Full}}  \\
 & \multicolumn{1}{l}{1-shot}  & 5-shot & \multicolumn{1}{l}{1-shot}   & 5-shot 
 \\ 
\shline
None (OTAM$^\dagger$)
&  82.4  & 91.1 
&  50.2 &  65.3 \\
\shline
BLIP (Default)
&  84.9  & 93.1 
&  51.9 &  68.2 \\
DeBERTa~\cite{DeBERTa}
&  84.4   &  92.8
&  \textbf{52.1}   &  \textbf{69.3} \\
CLIP~\cite{CLIP}
&  \textbf{85.8}   &  \textbf{93.5}
&   51.3  &  68.0 \\
\hline
\end{tabular}
\vspace{+1mm}
\caption{Ablation study on the effect of different text encoders.
}
\label{tab:text_encoder}
\vspace{-3mm}
\end{table}

\vspace{1mm}
\noindent \textbf{Impact of the quality of generated captions.}
Our experiments are all based on the officially released BLIP$_{\textrm{ViT-B}}$ model pretrained on 129M image-text pairs to generate captions.
In Table~\ref{tab:caption_method}, we investigate the effect of caption quality by varying the model size or the amount of pretraining data.
We can observe that larger models or more pretraining data usually lead to better caption generation, resulting in superior few-shot action recognition performance.

\begin{figure*}[t]
  \centering
   \includegraphics[width=0.98\linewidth]{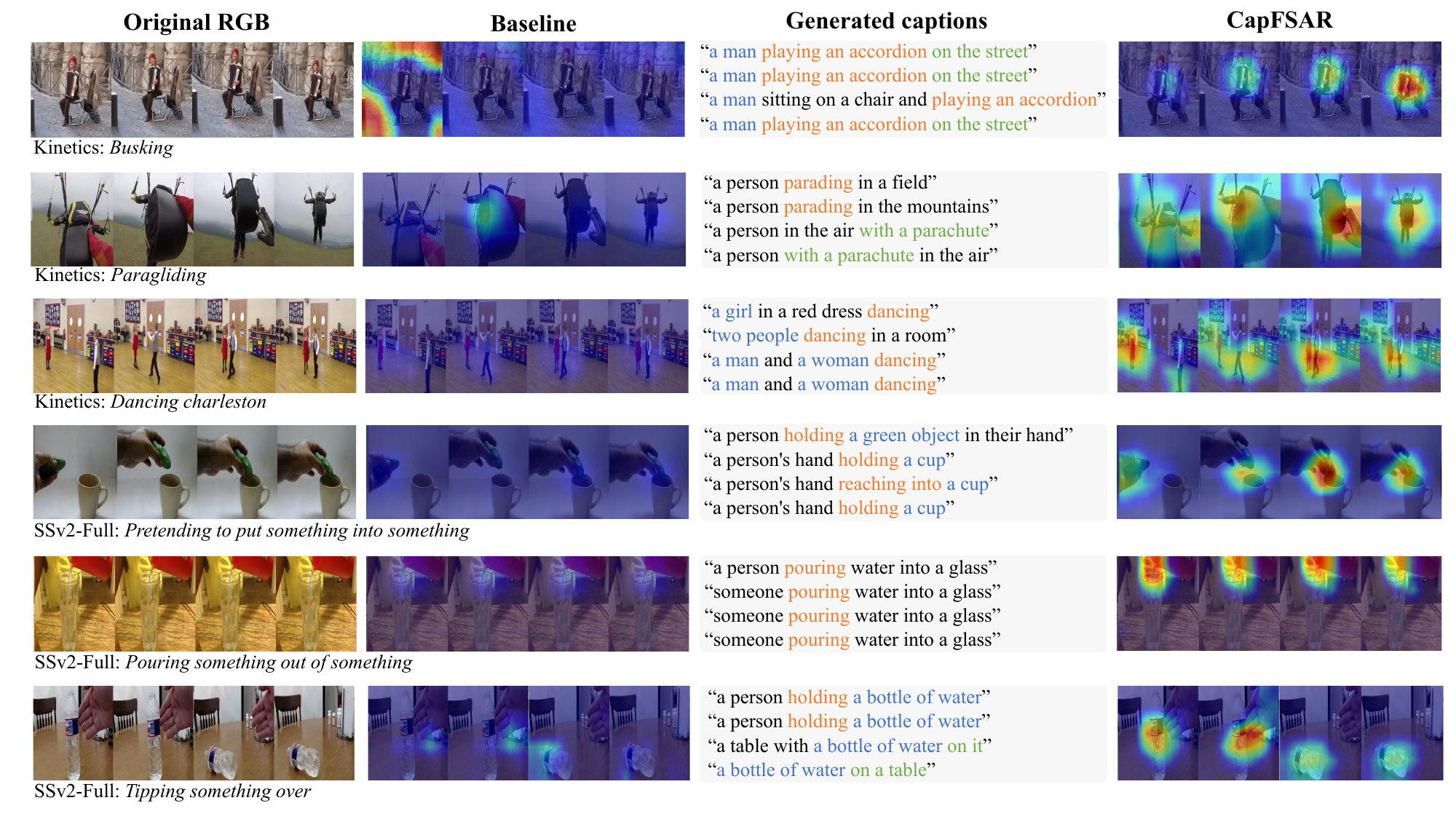}
\vspace{-1mm}
   \caption{
  Examples of the generated captions and GradCAM~\cite{selvaraju2017grad} heat maps on test sets of Kinetics (first three lines) and SSv2-Full (last three lines).
  For illustrative purposes, we
highlight words in \textcolor[RGB]{237,125,49}{orange} to represent actions and human-object interactions. 
Words in \textcolor[RGB]{68,114,196}{blue} reveal
subject or object, and the \textcolor[RGB]{112,173,71}{green} words indicate the scene.
The visual-only OTAM$^\dagger$ is leveraged as the baseline for comparison.
  %
}
   \label{fig:vis}
   \vspace{-3mm}
\end{figure*}

\begin{figure}[t]
  \centering
   \includegraphics[width=0.98\linewidth]{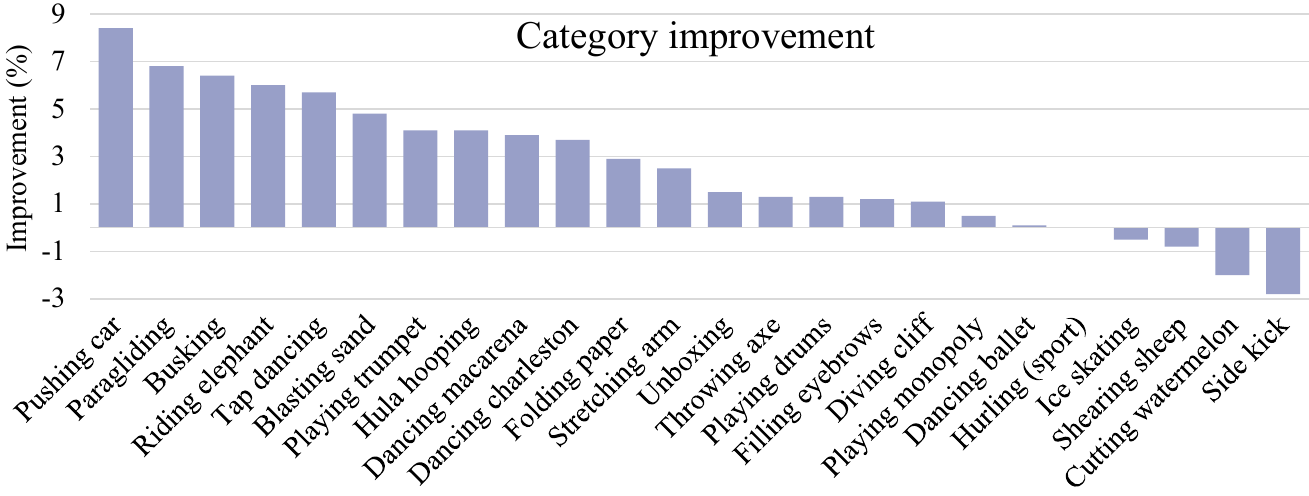}
\vspace{-1mm}
   \caption{
  5-way 1-shot class improvement of CapFSAR compared to the visual-only baseline OTAM$^\dagger$ on the Kinetics dataset.
}
   \label{fig:class_improvement}
   \vspace{-3mm}
\end{figure}

\vspace{1mm}
\noindent \textbf{Influence of text encoder types.}
For simplicity and convenience, we directly adopt the text encoder in the original BLIP~\cite{BLIP} model to encode caption representations.
To comprehensively explore the impact of different text encoders on performance, we leverage two widely used models, DeBERTa$_{\textrm{Large}}$~\cite{DeBERTa} and CLIP$_{\textrm{RN50x64}}$~\cite{CLIP}.
The results are presented in Table~\ref{tab:text_encoder}, and we can notice that different text encoders have specific bias differences.
Among them, CLIP performs best on the appearance-biased Kinetics dataset, and DeBERTa surpasses the other two on the motion-biased SSv2-Full dataset.
It is worth mentioning that the above variants all outperform the single visual modality baseline OTAM$^\dagger$, \eg, CLIP's 85.8\% 1-shot Kinetics result exceeds 82.4\% of OTAM$^\dagger$ by 3.4\%, revealing the generalizability of our framework.
Note that this observation also indicates the potential advantage of CapFSAR to exploit other advanced large language models in the future.
%

\section{Qualitative Analysis}
To analyze the role of text in our CapFSAR, we perform a qualitative study of the generated captions and gradient heat maps.
The visualization results are displayed in Figure~\ref{fig:vis}. 
We can observe that the auxiliary captions usually contain relevant information that can be leveraged to help extrapolate the correct classification results.
By comparing the heat maps of baseline and CapFSAR, we can clearly find that our CapFSAR focuses more on the discriminative regions, indicating the effectiveness of adding textual cues to assist in producing representative multimodal features.

\begin{figure}[t]
  \centering
   \includegraphics[width=0.95\linewidth]{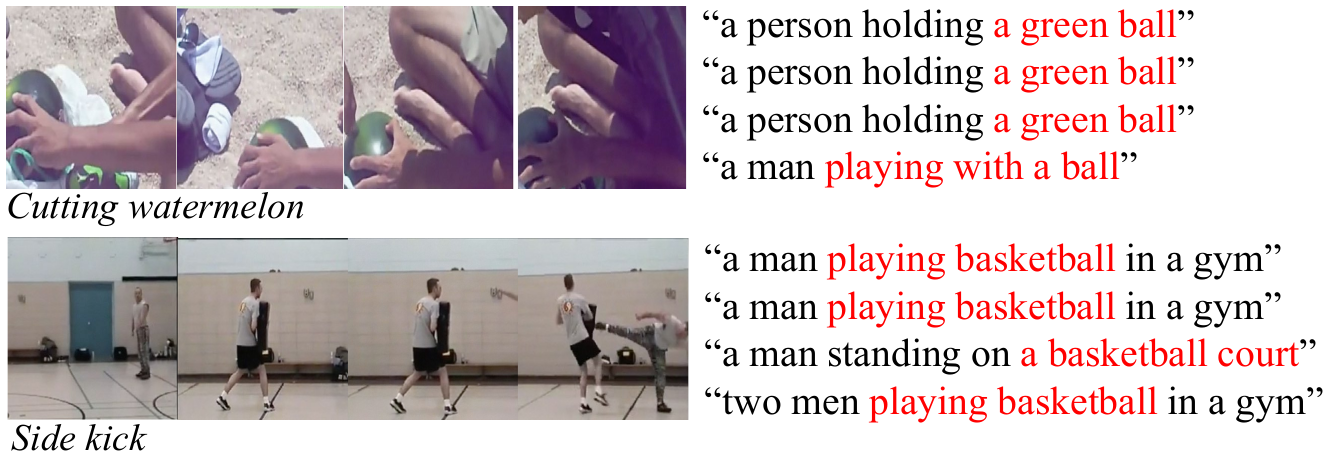}
\vspace{-1mm}
   \caption{
  Failure cases of generated captions from BLIP~\cite{BLIP}.
}
   \label{fig:fail_case}
   \vspace{-4mm}
\end{figure}

In Figure~\ref{fig:class_improvement}, we statistics on the category improvement of CapFSAR on the Kinetics dataset compared to the baseline OTAM$^\dagger$. It can be seen that there is a certain improvement in most action categories.
Some classes
see a significant improvement, \eg, ``Pushing car" and `` Paragliding", and we attribute this to the fact that the generated captions can easily include objects involved in these actions, such as ``car" and ``parachute".
In Figure~\ref{fig:fail_case}, we also present some failure cases.
We notice that due to some misleading appearances, such as ``watermelon looks like a green ball" and ``kicking a leg on a basketball court", the synthetic descriptions may be inaccurate and ultimately lead to wrong predictions.

\section{Limitations}
%
CapFSAR relies on captioning foundation models to generate high-quality captions and cannot be directly applied to traditional models with small pretraining data.
%
%
In addition, CapFSAR requires the generation of additional textual descriptions and thus will lead to increased inference costs. 
This can be alleviated by utilizing a lightweight pretrained caption decoder, which we leave for future work.

\section{Conclusion}

In this work, we presented a simple yet effective CapFSAR framework for few-shot action recognition.
%
%
CapFSAR succeeds in leveraging existing pretrained captioning foundation models to synthesize high-quality captions and thus help to obtain discriminative multimodal features for classification.
%
Extensive experiments on multiple benchmarks demonstrate that  CapFSAR outperforms existing  baselines and achieves state-of-the-art results under various settings.


\vspace{2mm}
\noindent{\textbf{Acknowledgements.}}
This work is supported by the National Natural Science Foundation
of China under grant U22B2053 and Alibaba Group through Alibaba Research Intern Program.

{\small
\bibliographystyle{ieee_fullname}
\bibliography{egbib}
}

\end{document}